\documentclass[letterpaper]{article}

\usepackage{aaai24} 

\usepackage{times}   

\usepackage{helvet}

\usepackage{courier}

\usepackage[hyphens]{url}

\usepackage{graphicx}

\usepackage{natbib}

\usepackage{caption} 
\usepackage{listings}

\usepackage{amsfonts}
\usepackage{amsmath}
\usepackage{booktabs}
\usepackage{newfloat}

\title{MABViT - Modified Attention Block Enhances Vision Transformers }
\author {
    Mahesh Ramesh\textsuperscript{\rm 1}\equalcontrib,
    Aswinkumar Ramkumar\textsuperscript{\rm 2}\equalcontrib
}
\affiliations {
    \textsuperscript{\rm 1}Indian Institute of Technology Madras\thanks{Work done while at Indian Institute of Technology Madras}\\
    \textsuperscript{\rm 2}NVIDIA\\
    mram18900@gmail.com, aswkumar@nvidia.com
}

\usepackage{bibentry}
\begin{document}

\maketitle

\begin{abstract}

Recent studies have demonstrated the effectiveness of Gated Linear Units (GLU) in enhancing transformer models, particularly in Large Language Models (LLMs). Additionally, utilizing a parallel configuration within each Transformer block rather than the conventional serialized method has been revealed to accelerate the training of LLMs without significantly impacting performance. However, when the MLP and attention block were run in parallel for the image classification task, we observed a noticeable decline in performance. We propose a novel transformer variant that integrates non-linearity within the attention block to tackle this problem. We implemented the GLU-based activation function on the Value tensor, and this new technique surpasses the current state-of-the-art S/16 variant of Vision Transformers by 0.6\% on the ImageNet-1K dataset while utilizing fewer parameters. It also supersedes the B/16 variant while using only half the parameters. Furthermore, we provide results with the GELU activation function variant to confirm our assertions. Lastly, we showcase that the MABViT variants exhibit greater potential when utilized in deep transformers compared to the standard architecture.

\end{abstract}

\section{Introduction}

The Transformer model \cite{c:23} is a widely adopted neural network architecture across multiple domains, including machine translation, image classification, and speech synthesis. The parallel configuration \cite{c:24} within each Transformer block rather than the conventional sequential structure has been revealed to accelerate the training of Large Language models (LLMs) by 15\% \cite{c:25} without significantly compromising the results. This prompted our investigation into its application in Vision Transformers.  

However, a disparity emerges between the parallel and standard formulations when training vision models, potentially attributed to the differences in scale between vision models (ranging from 5M to 100M parameters) and the considerably larger size of Language Models (greater than 1B parameters). Palm \cite{c:25} noted a minor decline in quality at the 8B scale with the parallel formulation but observed no such impact at the 62B scale compared to the standard Transformer. This observation regarding the success and limitations of the parallel formulations in models of varying scales motivated our exploration into integrating non-linearity within the attention block.

In this work, we aimed to identify the underlying reasons behind the comparable performance of parallel and standard structures at a large scale. Building on this understanding, we have developed a novel architecture that surpasses the performance of traditional transformer structures, achieving superior results with fewer parameters. 

The key contributions outlined in this paper are as follows:

\begin{itemize}
\item To the best of our knowledge, we are the first to analyze the difference in the performance of parallel structures at different scales. 
\item We hypothesize that representation collapse is the cause for the similarity in performance of parallel and standard transformer architectures and experimentally attempt to verify this claim.
\item We incorporate GLU-based activation functions within the attention block of the Transformer architecture to partially overcome the representation collapse issue.  
\item We demonstrate that MABViT-GLU variants outperform the standard architectures with fewer parameters.
\item We also provide results for the MABViT-GELU variant to reinforce our assertion that applying an activation function to the Value tensor within the attention module enhances the Vision Transformer. 
\item Finally, we exhibit that MABViT variants possess greater potential in deep transformers compared to the standard architecture.
\end{itemize}

\section{Background}

\subsection{Transformers}

The Transformer architecture is comprised of two key sub-modules: 

1) \textbf{Multi-Head Attention Module}: This module facilitates focusing on different positions within the input sequence to compute representations. It involves splitting the input into multiple heads to perform parallel self-attention before combining the outcomes.

2) \textbf{MLP Module}: Also known as the position-wise feed-forward network, this module processes the output from the attention mechanism through a series of fully connected layers independently at each position in the sequence.

\subsubsection{Pre-LayerNormalization Transformer}

The computation process in the Pre-LN transformer architecture can be represented as:
\begin{align}
X &= X + \text{MultiheadAttention}(\text{LN}(X)) \\
X &= X + \text{MLP}(\text{LN}(X))
\end{align}

Here, \(X\) denotes the input, and the operations involving the Multi-Head Attention and MLP (Multi-Layer Perceptron) modules are executed successively within the Transformer block. 

\subsubsection{Post-LayerNormalization Transformer}

The computations in the Post-LN transformer architecture are: 
\begin{align}
X &= \text{LN}(X + \text{Multiheadattention}(X)) \\
X &= \text{LN}(X + \text{MLP}(X))
\end{align}

Note: LN represents LayerNorm, and there is a difference in its position in the two architectures.

\subsection{Vision Transformers}

In Vision Transformers (ViT) \cite{c:26}, the architecture begins with a Patch Embedding (PE) layer that restructures the image into a sequence of patches. The PE layer first rearranges the image, denoted as $X \in \mathbb{R}^{H \times W \times 3}$, into patches $X_p \in \mathbb{R}^{\frac{HW}{P^2} \times P^2}$, where $P$ determines the patch dimensions. Each of these patches then undergoes an independent dense transformation, generating the visual tokens $X_t \in \mathbb{R}^{\frac{HW}{P^2} \times D}$. After this layer, a series of Transformer blocks are applied, operating with self-attention, feed-forward layers, and residual connections similar to typical Transformer architectures. 

\subsection{Parallel Structure}

The computation process in the Parallel Pre-LN transformer architecture is:

\[X = X + \text{Multiheadattention}(\text{LN}(X)) + \text{MLP}(\text{LN}(X))\]

Here, the operations involving the Multi-Head Attention and MLP (Multi-Layer Perceptron) modules are performed in parallel within the Transformer block instead of sequentially. 

\subsection{Representational Collapse}

The challenge with the representation capability of Pre-LN transformers was first identified by Admin \cite{c:28}. As the number of layers increases, the $X$ term grows, making the output from the Multi-Head Attention, or MLP blocks, relatively insignificant.

\[
X = X + \text{Multiheadattention}(\text{LN}(X))
\]

For instance, here, as we move to deeper layers, the magnitude and variance of the $X$ term greatly exceeds the output of the Multi-Head Attention block. This suggests that the input and output values in the later blocks are likely to converge or become increasingly similar. 

\subsection{Gated Linear Units}

In their work, \cite{c:32} presented Gated Linear Units (GLU), a neural network layer created by combining two linear transformations of the input using element-wise multiplication. One of these transformations employs a sigmoid activation function.

\[
\text{GLU}(X, W, V, b, c) = \sigma(XW + b) \odot (XV + c)  
\]

This formulation for the GLU equation has the input $X$, weight matrices $W$ and $V$, bias vectors $b$ and $c$, utilizing the sigmoid function $\sigma$ and element-wise multiplication $\odot$.

\cite{c:31} introduced further variants of the GLU and demonstrated that substituting the initial linear transformation and activation function of the MLP layer in the transformer architecture with GLU or one of its variants enhances the Transformer's performance.
\[\text{MLP}_{\text{GEGLU}}(X, W_1, V, W_2) = ( \text{GELU}(XW_1) \odot XV )W_2\]
\[\text{MLP}_{\text{SwiGLU}}(X, W_1, V, W_2) = ( \text{Swish1}(XW_1) \odot XV )W_2\]

Each of these layers involves 3 matrices instead of 2, so the hidden dimension was reduced by 2/3 to maintain number of parameters.

\section{Related Work}

Several approaches have been proposed to overcome the representation collapse issue in Pre-LN transformers. Techniques like Admin \cite{c:28} and DeepNet \cite{c:29} add different weights to the residuals. Furthermore, 
DeepNet \cite{c:29} also modified the initialization to reduce training instability in Post-LN transformers. 
Few others, like Resi-Dual \cite{c:27}, proposed modifications to architecture to address this problem.

However, when applied to Vision Transformers, all these techniques either result in training instability or exhibit considerably inferior performance compared to standard Pre-LN transformer architectures. This motivated us to develop a novel architecture that addresses the representation collapse problem and achieves improved results on vision tasks over standard Vision Transformers.

\section{Methodology}

We hypothesize that representation collapse is the underlying cause for the comparable performance of parallel and standard transformer architectures at a large scale.

\subsubsection{Standard Pre-LN Computation}

\begin{align*}
X_{i+1} &= X_i + \text{Multiheadattention}(\text{LN}(X_i)) \\
X_{i+2} &= X_{i+1} + \text{MLP}(\text{LN}(X_{i+1}))
\end{align*}

\subsubsection{Parallel Pre-LN Computation}

\[
X_{j+1} = X_j + \text{Multiheadattention}(\text{LN}(X_j)) + \text{MLP}(\text{LN}(X_j))
\]

Considering an initial 1x1 tensor as input and positive outputs from both the Multihead-Attention and MLP blocks at each layer, as we progress through deeper layers: 

\[X_i\gg\text{Multiheadattention}(\text{LN}(X_i)) \rightarrow(X_{i+1} \approx X_i)\]

If \(X_{i} \approx X_{j}\), it implies that \(X_{i+2} \approx X_{j+1}\), supporting our hypothesis that the growth of the \(X\) term causes similarity in performance between parallel and standard transformer formulations under certain cases.

In Post-LN transformers, the normalization of the $X$ term takes place after every Multihead-attention and MLP block, a practice associated with training instability. Admin \cite{c:28} noted that when Post-LN transformers do not diverge, they tend to outperform the Pre-LN variant in machine translation. However, in the context of most vision tasks, Post-LN transformers often exhibit divergence, motivating us to develop a new transformer variant aimed at surpassing the performance of traditional ViTs.

\textbf{Objective}: Construct a novel architecture with equal or fewer parameters that surpass standard Pre-LN Vision Transformers

In Pre-LN transformers, we hypothesize that incorporating the Multi-Head Attention output into the residual before feeding it to the MLP block becomes redundant in deeper layers. To address this, we propose integrating the activation function within the attention module to apply a non-linear transformation to the Multi-Head Attention output effectively. 

To facilitate a non-linear transformation on the Multi-Head Attention output, we introduced an activation function on the Value tensor inside the attention module. Since the GLU activation function \cite{c:31} has proven effective in enhancing Transformer models, we opted to incorporate it as the activation in the attention block. We also conducted experiments with the GELU activation and compared it to the standard Vision Transformers.

\subsection{Standard Attention Block}

The computations inside a regular attention block are: 

\[
\begin{array}{ll}
Q = & XW_Q + B_Q \\  
K = & XW_K + B_K \\
V = & XW_V + B_V
\end{array}
\]

Given the input features or embeddings $X$, these are transformed into the $Q$, $K$, and $V$ matrices via learned linear projections. Here, $W_Q$, $W_K$, and $W_V$ represent the weight matrices while $B_Q$, $B_K$, and $B_V$ denote the bias terms for the linear projections.

Equations for computing the attention scores ($A$), attention distribution ($D$), and weighted values ($WV$):

\[
\begin{array}{ll} 
A = & \frac{QK^T}{\sqrt{d_k}} \\
D = & \text{softmax}(A) \\  
WV = & D \times V
\end{array}
\]

The attention scores ($A$) are determined by taking the dot product of the query ($Q$) and key ($K$) matrices. To scale these attention scores, we divide by the square root of the dimensionality $d_k$.
Subsequently, the softmax function is applied to these scores to produce the attention distribution ($D$) that provides the weight or significance assigned to each token in the input sequence. $D$ is then multiplied with the value matrix ($V$) to obtain the weighted values ($WV$) capturing the importance of the individual elements.

\begin{figure}[ht]
\centering
\includegraphics[width=0.17\textwidth]{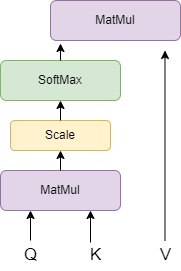} 
\caption{Scaled Dot Product Attention}
\label{fig1}
\end{figure}

\subsection{Modified Attention Block}

Similar to the standard attention block, we transform input into the Query, Key and Value. Following this, we apply the activation function to the Value tensor. Otherwise, the standard processes within the attention block remain unchanged.

\[
\begin{array}{ll} 
V = & XW_V + B_V \\
V = & Activation(V)  
\end{array}
\]

\begin{figure}[ht]
\centering
\includegraphics[width=0.2\textwidth]{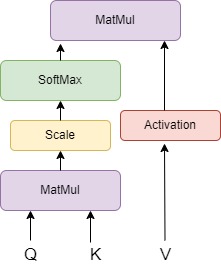} 
\caption{Modified Scaled Dot Product Attention}
\label{fig2}
\end{figure}

\subsection{GLU Variant}

For the GLU-based activation function, we increased the dimension of the value tensor to twice its original size, dividing it into two halves. We applied the activation to one half and performed element-wise multiplication with the other half:

\[
\text{V}_{\text{SwiGLU}}(X, W_1, W_2) = ( \text{Swish}(XW_1) \odot XW_2 )
\]

To counterbalance the additional parameters introduced by using the GLU activation, we reduced the number of parameters in the MLP block.

While the proposed architecture does not provide an exhaustive solution to the representation collapse problem, it partially addresses it by giving significance to the output of the Multi-head Attention block. 

\section{Experiments}

\subsection{Setup}

We utilized the Vision Transformers (ViT) model \cite{c:26}, which has demonstrated impressive performance across diverse visual tasks. Seven variants of ViT were trained on three different architectures (Ti/16, S/16, and B/16):

1) \textbf{Standard ViT}: We implemented the standard Vision Transformer using big-vision \cite{c:34}.

2) \textbf{Standard Parallel Structure}: We modified the standard architecture to perform Multi-Head Attention and MLP computations in parallel instead of sequentially.

3) \textbf{GLU-based variant}: We applied the GLU-based activation within the standard ViT attention module on the Value tensor without any additional hyperparameter tuning.

4) \textbf{GLU Parallel Variant}: Same as above but applied to the Parallel variant. 

5) \textbf{Parameter Reduced GLU Variant}: To compensate for the extra parameters from GLU, we reduced MLP dimensions from 4x to 3x embedding size.

6) \textbf{Parameter Reduced GLU Parallel Variant}: As above but on the Parallel ViT.

7) \textbf{GELU Variant}: GELU activation applied to standard ViT Value tensor.

The training was conducted following the AugReg methodology \cite{c:33} for 300 epochs with a batch size of 4096 for Ti/16 and S/16 architectures and 2048 for B/16. We evaluated the top-1 accuracy on the ImageNet-1K validation set per common practices. A dropout of 0.1 was implemented for the B/16 architecture, while other architectures were trained without dropout. The training process encompassed the standard ViT, Parallel ViT, standard and Parallel variants of the GLU-based MABViT architecture (with and without reducing MLP parameters), and the GELU-based MABViT architecture to provide a comprehensive and detailed analysis.

\subsection{Architectures}

\begin{center}
\begin{tabular}{ccccc} 
\toprule
Architecture & Layers & MHA Dim & MLP Dim & Heads \\
\midrule   
Ti/16 & 12 & 192 & 768 (576) & 3\\
S/16 & 12 & 384 & 1536 (1152) & 6\\
B/16 & 12 & 768 & 3072 (2304) & 12\\
\bottomrule
\end{tabular}
\end{center}

Note: The MLP dimensions in brackets correspond to the parameter-reduced variants.

\subsection{Number of Parameters} 

\begin{center}
\begin{tabular}{cccccc}
\toprule
Arch & GLU & PR-GLU & Base and GELU & \\
\midrule
Ti/16 & 6.2 M  & 5.3 M & 5.8 M \\
S/16 &  24 M & 20.4 M & 22.2 M \\
B/16 &  94 M &  80 M  & 86 M \\
\bottomrule
\end{tabular}
\end{center}
\leavevmode

For each architecture, the standard ViT and Parallel ViT pairs possess identical parameters. 

\subsection{Comparison Between Standard and Parallel Variants}

\begin{center}
\begin{tabular}{ccc} 
\toprule
Architecture & Base & Parallel \\
\midrule
Ti/16 & 72.444 & 70.99 \\
S/16 & 78.832 & 77.76 \\
B/16 & 80.174 & 79.824 \\  
\bottomrule
\end{tabular}
\end{center}

There is a noticeable difference between the validation accuracy of the standard and parallel architectures across all the models. With smaller-scale models, the magnitude and variance of the \(X\) term are reduced. Consequently, the standard architectures attain superior performance compared to the parallel formulations. We observe that the difference in results reduces when we increase the width of the layers. We hypothesize that the magnitude of the $X$ term rises with greater width, causing the parallel and standard structures to perform more similarly.

\subsection{Comparison Between Standard and Parallel GLU variants}

\begin{center}
\begin{tabular}{ccc}  
\toprule
Architecture & GLU-Base & GLU-Parallel \\
\midrule 
Ti/16 & 73.292 & 72.25 \\ 
S/16 & 79.956 & 78.85 \\
B/16 & 78.35 & 79.236 \\
\bottomrule
\end{tabular}
\end{center}

The standard and parallel GLU-based variants demonstrate a steady enhancement of over 1\% compared to their conventional counterparts on the S/16 and Ti/16 architectures. However, on the B/16 architecture, both variants exhibit overfitting and underperformance. 
As expected, the parallel GLU variant surpasses the standard GLU on B/16 since it is less prone to overfitting.

\subsection{Comparison Between Parameter Reduced GLU and Parameter Reduced Parallel GLU variants}

\begin{center}
\begin{tabular}{ccc}
\toprule
Architecture & GLU-PR-Base & GLU-PR-Parallel \\  
\midrule
Ti/16 & 72.604 & 71.67 \\
S/16 & 79.44 & 78.33 \\
B/16 & 78.448 & 79.37 \\
\bottomrule
\end{tabular}
\end{center}

As described earlier in the architecture specifications, the GLU-PR-base and GLU-PR-Parallel variants possess fewer parameters compared to the standard ViT architecture. Both the PR-GLU S/16 and Ti/16 variants surpass their traditional counterparts. The PR-GLU base variant exhibits a 0.6 \% improvement over the standard ViT on S/16. However, as with the previous GLU variants, overfitting occurs again on the B/16 architecture for both formulations.

\subsection{GELU Variant}

\begin{center}
\begin{tabular}{ccc} 
\toprule
Architecture & PR-GLU-Base & GELU-Base \\
\midrule
Ti/16 & 72.604 & 72.598 \\
S/16 & 79.44 & 79.37 \\
\bottomrule  
\end{tabular}
\end{center}

Our experiments with the GELU variants reveal that the GELU S/16 and Ti/16 variants outperform the standard ViT but fall short compared to the GLU-base and PR-GLU-base variants. This reaffirms our assertion that integrating an activation function within the attention module enhances the Vision Transformer's performance. 

\subsection{Experiments on M/16 Architecture}

Since the B/16 MABViT variants exhibited overfitting, we conducted additional experiments on an intermediate architecture between S/16 and B/16. The M/16 architecture has 12 layers, 576 dimensions in the MHA, 2304 dimensions in MLP and consists of 8 heads. 

We evaluated all four standard (not parallel) variants with the same hyperparameters.  

\begin{center}
\begin{tabular}{ccc}
\toprule
Variant & Parameters & Accuracy \\   
\midrule
Base & 39.1 M & 79.752 \\
GELU-Base & 39.1 M & 80.256 \\
GLU-Base & 42.25 M & 80.27 \\ 
PR-GLU-Base & 35.9 M & 80.324 \\
\bottomrule
\end{tabular}
\end{center}

All the M/16 MABViT variants surpass the baseline architecture. Furthermore, they outperform the base B/16 variant while utilizing only half the number of parameters. This demonstrates the ability of MABViT variants to capture complex patterns with fewer parameters efficiently.

\subsection{Experiments on S/16 with 18 Layers}

In our final experiments, we increased the number of layers in the S/16 architecture to 18 and evaluated the standard variants:

\begin{center}
\begin{tabular}{cc} 
\toprule
Variant & Accuracy \\
\midrule
Base & 78.466 \\  
GELU & 80.048 \\
GLU-Base & 80.214  \\
PR-GLU-Base & 80.262 \\
\bottomrule
\end{tabular}
\end{center} 

As anticipated, with greater depth, the increasing magnitude of the $X$ term impedes the performance of the baseline model. However, the MABViT variants are able to partially overcome the representation collapse issue by providing significance to the output of the MHA block. Evidently, the MABViT variant continues to improve as we raise the number of layers.

Overall, applying an activation function to the Value tensor boosts the performance of the Ti/16, S/16 and M/16 variants for both standard and parallel formulations. The overfitting exhibited by the GLU B/16 variants indicates the ability of the new architecture to capture complex patterns present in the dataset. The results also accentuate the importance of the standard structure for smaller-scale models. 

\section{Discussion} 

Although the suggested modification provides significance to the Multi-Head Attention block's output, the representational collapse problem is not entirely resolved. The linear growth of the $X$ term continues, resulting in convergence between the input and output of deeper layers. However, the proposed architecture improves performance in initial layers and assigns greater weight to the Multi-Head Attention output in later layers compared to the standard ViT. Additional research could provide potential solutions to tackle representation collapse fully or refine this technique.

\begin{figure}[ht]
\centering
\includegraphics[width=0.4\textwidth]{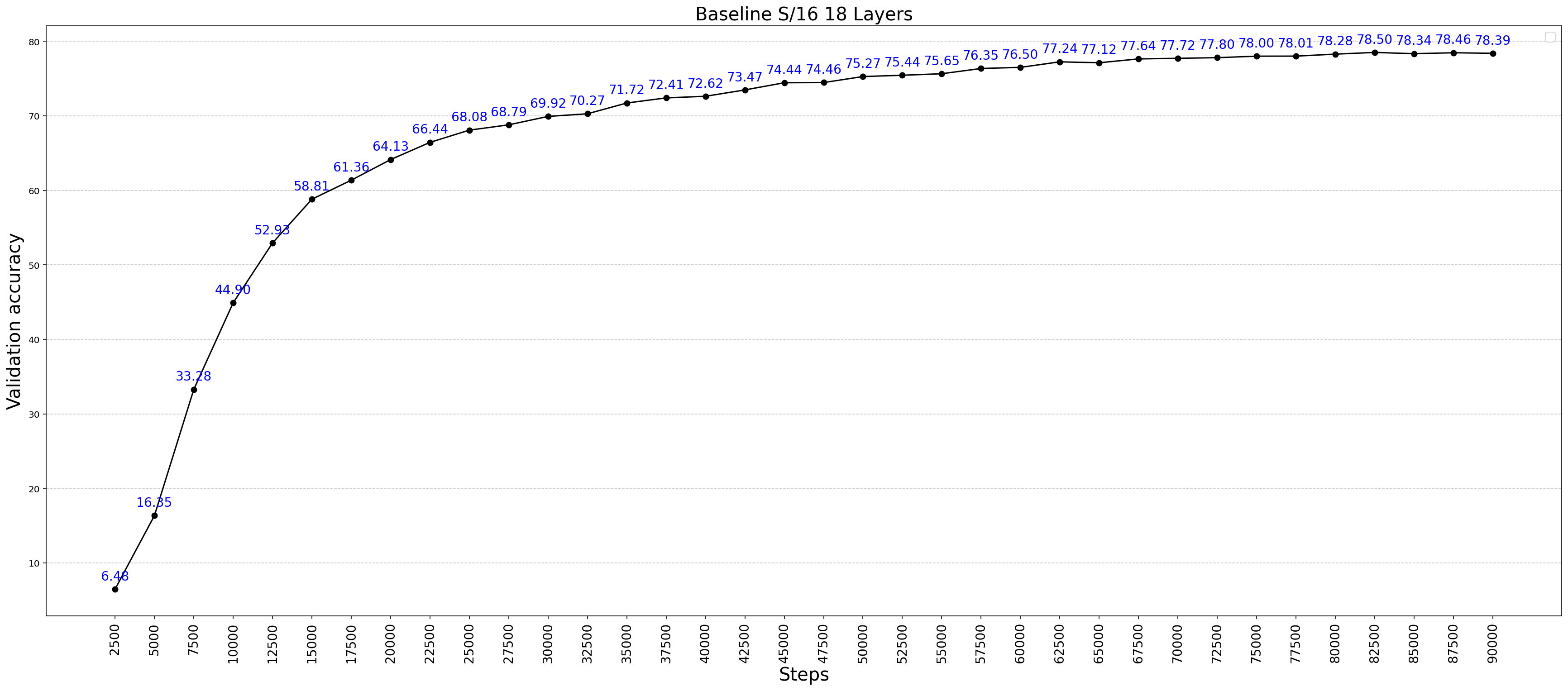} 
\caption{Validation accuracy progression of the Baseline S/16 18 Layers variant over 90,000 training steps.}
\label{fig3}
\end{figure}
\begin{figure}[ht]
\centering
\includegraphics[width=0.4\textwidth]{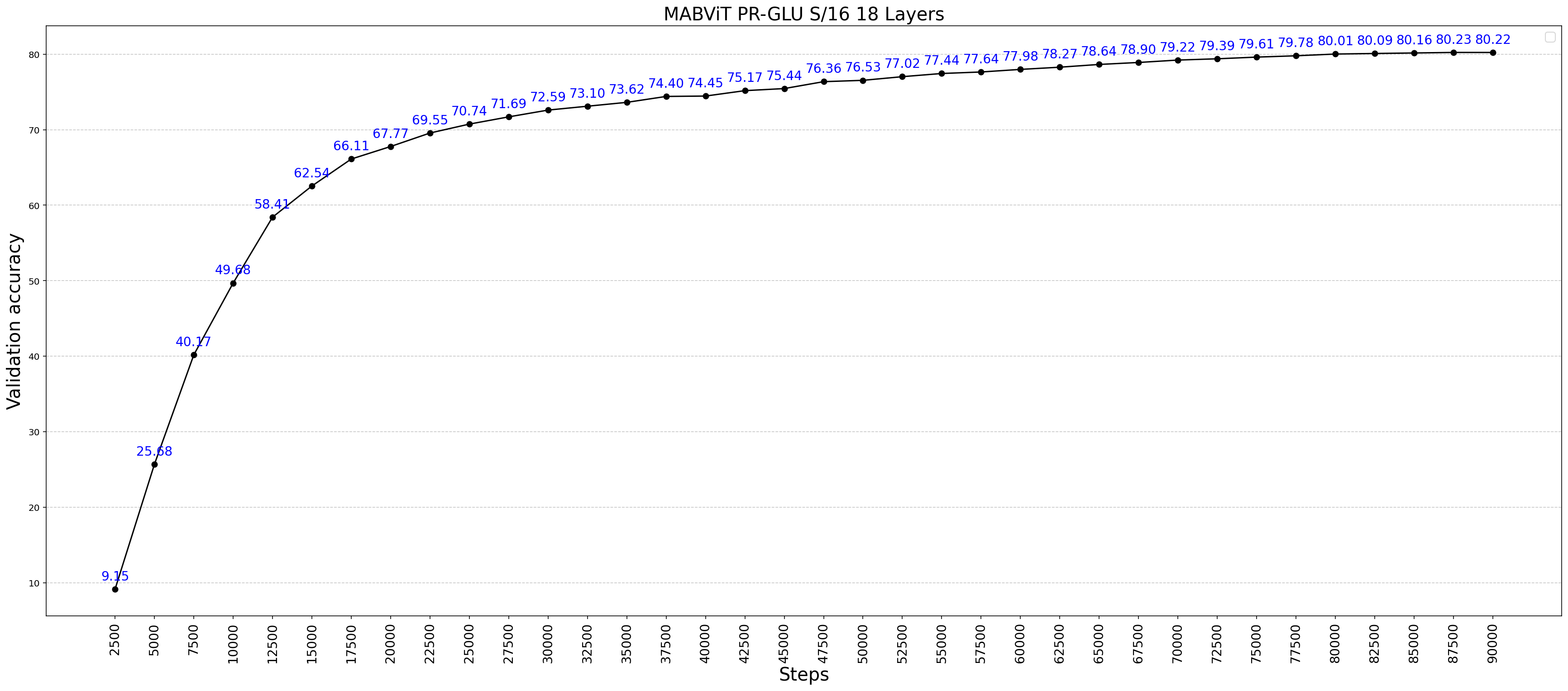} 
\caption{Validation accuracy trajectory of the MABViT PR-GLU S/16 18 Layers variant over 90,000 training steps.}
\label{fig4}
\end{figure}\begin{figure}[ht]
\centering
\includegraphics[width=0.4\textwidth]{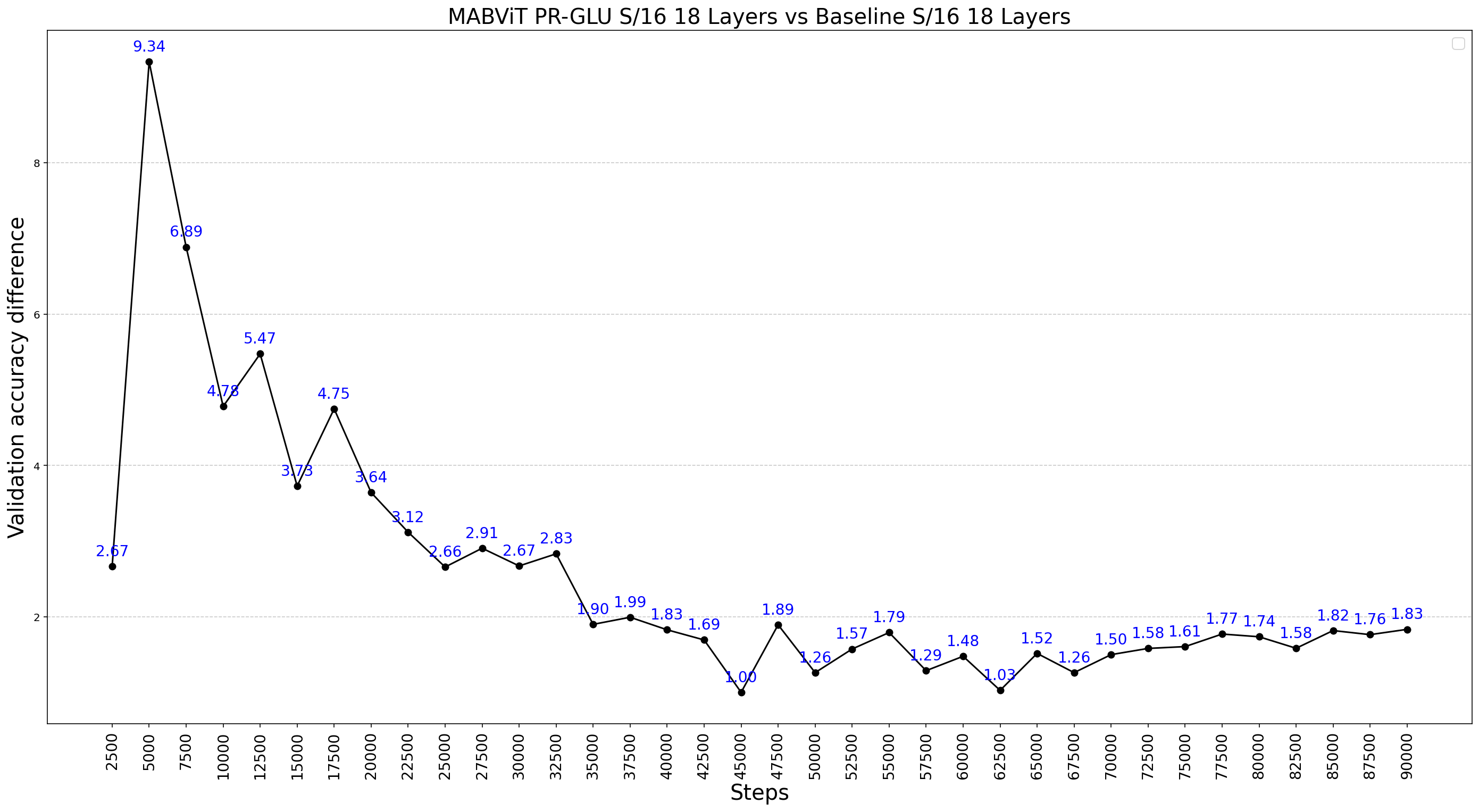} 
\caption{Difference between validation accuracy of MABViT PR-GLU-Base S/16 18L vs Base S/16 18L over 90,000 training steps}
\label{fig5}
\end{figure}
Our experiments demonstrate that the MABViT variants converge substantially faster than the standard architectures. From Figure \ref{fig3} and Figure \ref{fig4}, we can see that the MABViT PR-GLU variant achieves 78 \% validation accuracy at the $62500^{th}$ step, whereas baseline ViT attains 78 \% at the $75000^{th}$ step. Across all experiments, the MABViT models consistently exhibit superior performance throughout training, barring cases of overfitting. The new model illustrates an ability to swiftly recognize complex patterns in the dataset compared to the standard variant, which is especially relevant as most Large Language Models are trained on massive data. Our work also emphasizes the significance of the Value tensor projection layer, motivating prospective research into utilizing it as a Mixture of Experts.

Note: The validation accuracy reported in the tables corresponds to the conclusion of 300 epochs (93,000 steps). It is important to note that the figures were generated for 90,000 steps for clearer visualization.

\section{Conclusion}

In this work, we successfully show that representation collapse causes the comparable performance of parallel and standard transformer architectures at scale. We also demonstrate that current transformer architectures can be improved through partially resolving representational collapse by effectively integrating non-linearity inside the attention module using GLU variants. The PR-SwiGLU S/16 variant enhances performance by 0.6\% with fewer parameters, and all the MABViT M/16 variants surpass the standard B/16 architecture, utilizing only half the parameters. Additionally, we provided analysis with GELU activation and substantiated that activation inside the attention module benefits Vision Transformer. Furthermore, we exhibited that MABViT variants possess greater potential in deep transformers compared to the standard architectures.

\section{Acknowledgement}
We extend our sincere thanks to Bharat Kumar Sharma and NVIDIA for their crucial support in the project.

\bibentry{c:35}
\bibentry{c:36}
\bibentry{c:37}
\bibentry{c:38}
\bibentry{c:30}
\bibentry{c:39}
\bibentry{c:41}
\bibentry{c:40}
\bibentry{c:42}
\bibentry{c:44}
\bibliography{aaai24}

\end{document}